\documentclass[runningheads]{llncs}

\usepackage{eccv}
\usepackage{eccvabbrv}

\usepackage{graphicx}
\usepackage{booktabs}
\usepackage{siunitx}
\usepackage[table]{xcolor}
\usepackage{multirow}
\usepackage{makecell}
\usepackage{adjustbox} 
\usepackage{pifont}

\usepackage[accsupp]{axessibility}  

\usepackage{hyperref}
\usepackage{orcidlink}

\usepackage{acronym}
\acrodef{DATASET}{NVSign}
\acrodef{PHOENIX}{Phoenix14T}
\acrodef{BSLCP}{BSL Corpus}

\begin{document}

\title{SignRefine: Adapting Foundational Video Models for Sign Language Generation}
\titlerunning{SignRefine}

\author{Anton Pelykh\inst{1}\orcidlink{0009-0005-9075-5718} \and
Edward Fish\inst{1}\orcidlink{0009-0004-2964-8430} \and
Ozge Mercanoglu Sincan\inst{1}\orcidlink{0000-0001-9131-0634} \and
Richard Bowden\inst{1}\orcidlink{0000-0003-3285-8020}}

\authorrunning{A.~Pelykh et al.}

\institute{Centre for Vision, Speech and Signal Processing, University of Surrey, Guildford, UK \\
\email{\{a.pelykh,edward.fish,o.mercanoglusincan,r.bowden\}@surrey.ac.uk} \\
\url{https://cogvis-cvssp.github.io/papers/signrefine/}}

\maketitle

\begin{abstract}
Sign language video generation demands precise hand and facial articulation, yet modern video diffusion models, trained predominantly on spoken-language video, produce artifacts that render signing unintelligible. We propose \textbf{SignRefine}, a sign language video generation model that produces comprehensible signing from 2D keypoint conditioning alone, generalizing across appearances and visual conditions. Our approach builds on a pretrained video diffusion transformer and introduces local adapters with spatial grounding to selectively refine hand and face regions, steering the strong base model’s prior toward accurate articulation. To enable this work and support broader sign language research, we present \textbf{\acl{DATASET}}, a large-scale dataset of video content natively produced in sign language, offering diverse signer appearances, environments, and natural conversational settings. Trained on this data, our model shows up to 30\% improvement in hand pose precision metrics over the strongest baseline and is preferred by sign language users for visual quality and comprehensibility in more than 80\% of comparisons.
\end{abstract}
\section{Introduction}
\label{sec:intro}
\begin{figure}[t]
    \centering
    \includegraphics[width=\linewidth]{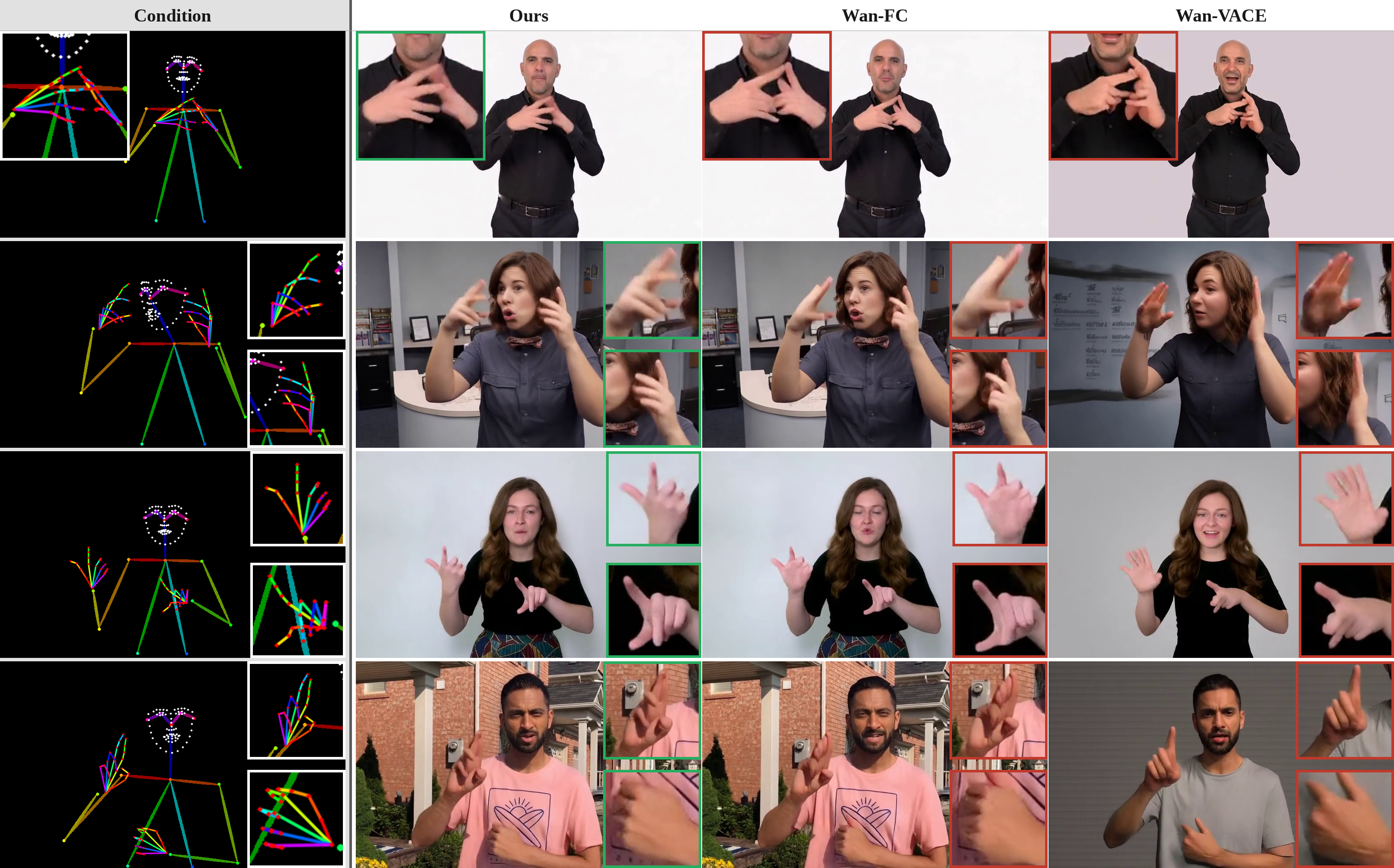}
    \caption{Qualitative comparison of sign language video generation approaches from 2D keypoint conditioning. From left to right: input keypoint condition, Our model, Wan2.1-1.3B-Fun-Control \cite{wang2025wan}, Wan2.1-1.3B-VACE \cite{jiang2025vace}.}
    \label{fig:gen_example_main}
\end{figure}

Sign languages are the primary means of communication for Deaf communities worldwide. Lacking a widely adopted written form, video is the natural medium for sign language content creation, accessibility, and communication. Consequently, there is a critical need for automatic sign language video generation. While modern Video Diffusion Models (VDMs) produce impressively realistic general-purpose video, they consistently fail to produce plausible sign language outputs (see Fig.~\ref{fig:gen_example_main}). This is because the complex articulation of hands and nuanced facial expressions required for sign language are largely out-of-distribution for the training data of these models. Furthermore, hands and face, which carry the core linguistic meaning, are fine-grained structures occupying a small fraction of the frame. In the patchified latent space where VDMs operate, these regions are aggressively compressed spatially and their structural details are overwhelmed by the global reconstruction objective. As a result, standard full-body conditioning cannot recover this lost definition, resulting in corrupted or under-articulated hand and face regions.

Currently, the field of research faces a stark duality between specialized and generalized video generation. Existing sign language-specific video models \cite{stoll2018sign, saunders2022signing, saunders2020everybody, shi2024pose, qi2024signgen, fang2025signdiff, wang2025advanced} are typically trained from scratch on small, constrained datasets. While they can produce plausible results on in-distribution data, they overfit to specific signer appearances and fail to generalize to new individuals, scales, or visual environments, producing restrictive, low-resolution outputs. This is amplified by the prevalence of interpreted content in existing datasets, which lacks the spatio-temporal complexity and fluidity of natural signing. In contrast, foundational human video models \cite{hu2024animate, xu2024magicanimate, zhu2024champ, wang2025multi, wang2024disco, wang2025unianimate, Wang2025UniAnimateDiTHI, gan2025humandit} showcase exceptional generalizability and visual quality but cannot handle the significant domain shift required for continuous signing. To date, no approach successfully resolves these issues to produce high-fidelity, generalizable sign language video that is actually comprehensible to Deaf viewers.

Modern sign language production approaches typically consist of two stages: (1) translating spoken language into a discrete intermediate representation such as skeletal data or latent motion tokens \cite{yu2024signavatars, wang2025advanced, saunders2020progressive, zuo2025signs}, (2) generating videos using these representations \cite{saunders2020everybody, saunders2022signing, wang2025advanced}. In this work, we tackle the second part of the pipeline -- generating anatomically precise and expressive sign language video across diverse appearances given the driving skeleton sequence. To achieve this, we adapt a powerful prior of a large-scale video Diffusion Transformer (DiT) \cite{peebles2023scalable} for sign language using our proposed local adapters with spatial grounding. Aligning with sign language linguistics, where manual and non-manual signals operate as distinct communicative channels, we isolate the conditioning signals for each region (face, left and right hands) and process them with separate spatial condition encoders at the increased resolution. To reinforce the spatial alignment of the extracted regional conditions, we also encode the locations of each region in the frame using the CoordConv approach \cite{liu2018intriguing}. The extracted regional features are then integrated into the DiT backbone with our local cross-attention adapters to selectively refine the quality of the hand and face areas. To spatially ground the refinements, we employ an attention masking strategy with learned sink tokens, ensuring that background queries are routed away from the foreground.

To address visual sterility and linguistic biases prevalent in existing sign language datasets, we introduce \acl{DATASET}, a large-scale dataset of native sign language content. The dataset provides a vast diversity of signer appearances, dynamic visual environments, and camera framings. Beyond its utility for the current work in sign language video generation, the dataset's significant volume of authentic, multi-participant conversation unlocks new research directions for the broader AI community, enabling the study of multi-signer dynamics, turn-taking, and complex non-manual discourse markers.

In summary, our contributions are as follows:
\begin{enumerate}
    \item We introduce the first generalizable pose-to-video generation approach designed to improve the comprehensibility of generated sign language content.
    \item We develop a novel conditioning scheme for DiT-based human video models using local adapters with spatial grounding. It enables targeted refinement of hand and face areas without degrading the quality and generalizability of the backbone model.
    \item We introduce \acl{DATASET}, a large-scale multi-modal natively signed dataset.
    \item Through rigorous quantitative evaluation and a user study with sign language users of different levels, we demonstrate that our approach outperforms the baselines and is capable of generating comprehensible and diverse sign language videos.
\end{enumerate}
\section{Related Work}
\label{sec:lit}
\textbf{Video Generation Models} ~ The field of video generation progressed from GAN-based \cite{goodfellow2020generative, vondrick2016generating, clark2019adversarial, tulyakov2018mocogan, saito2017temporal, wang2020g3an, skorokhodov2022stylegan} and traditional Likelihood-based \cite{kalchbrenner2017video, kumar2019videoflow, yan2021videogpt, denton2018stochastic, hong2023cogvideo} approaches to Video Diffusion Models (VDMs), which have established a new performance frontier in sample quality and scalability. Early VDMs directly extended the 2D DDPM framework \cite{ho2020denoising} to the time dimension in pixel space \cite{ho2022video} and later in the latent space of a pre-trained autoencoder \cite{blattmann2023align}. This shift to latent space drastically improved model efficiency and unlocked high-resolution generation. Instead of training from scratch, numerous works \cite{guo2024animatediff, blattmann2023stable, blattmann2023align, zhou2211magicvideo, singer2023makeavideo, wu2023tune, wang2023modelscope} expanded existing powerful image Latent Diffusion Models (LDM) \cite{rombach2022high} to the video domain, sparking a modern trend of adapting strong visual priors for downstream generative tasks.

Alongside generation quality, the research community has heavily focused on the controllability of VDMs. ControlNet-style adapters \cite{zhang2023adding} have been widely adopted to drive video generation using additional visual modalities \cite{zhang2024controlvideo, hu2023videocontrolnet, lin2025ctrladapter, peng2024controlnext, wang2024easycontrol} and camera controls \cite{yin2023dragnuwa, he2024cameractrl}. Recently, the paradigm has shifted away from the traditional denoising U-Net architecture \cite{ronneberger2015u} toward DiT \cite{peebles2023scalable}, which treats video latents as a sequence of spacetime patches and scales more predictably with data and compute. The latest DiT-based video models \cite{videoworldsimulators2024, ma2025latte, yang2025cogvideox, kong2024hunyuanvideo, genmo2024mochi, wang2025wan} generate longer, highly consistent videos with complex interactions. Crucially, control mechanisms originally designed for U-Nets can also be successfully adapted to these DiT architectures.

\noindent\textbf{Human Image Animation} ~ Advancements in LDMs directly catalyzed developments in human image animation. Several works \cite{hu2024animate, xu2024magicanimate, zhu2024champ, wang2025multi} utilize a frozen LDM as a feature extractor for a reference image. These features are then injected into the backbone video model to produce an animated human video. Other approaches \cite{wang2024disco, wang2025unianimate} employ distinct encoding and fusion mechanisms for reference images and driving keypoint sequences before passing them into the backbone diffusion model. Progressing alongside foundation models, recent frameworks \cite{Wang2025UniAnimateDiTHI, gan2025humandit, cheng2025wan} have adopted DiTs as their backbone. However, while these models excel at general human movement and global pose alignment, they struggle to reliably produce highly articulated hand shapes and nuanced facial expressions crucial for complex communication in sign languages.

\noindent\textbf{Sign Language Video Generation} ~ Given the unique anatomical demands of continuous signing, the generation of sign language videos has historically developed as a specialized domain. For several years, pose-conditioned GAN approaches \cite{stoll2018sign, saunders2022signing, saunders2020everybody} served as the foundational standard for this task, successfully establishing the viability of generating continuous sign language videos. While pioneering, these early models typically overfit to specific appearances, lacked flexible visual conditioning, and struggled to generalise across varying body scales and rotations. Recent works \cite{shi2024pose, qi2024signgen, fang2025signdiff, wang2025advanced} have adopted diffusion models to generate more temporally consistent and expressive sign videos. However, because these models are trained from scratch on small domain-specific datasets, they produce low-resolution outputs and suffer from poor generalisation. Furthermore, some approaches \cite{fang2025signdiff, wang2025advanced} rely on additional dense modalities such as edge maps or posed 3D hand meshes, limiting their practical applicability. In contrast, we adapt a large-scale pretrained video diffusion model, inheriting its strong visual priors and generalisability. Moreover, our method is conditioned on sparse skeleton keypoints, avoiding dense structural modalities to ensure broad real-world utility.
\section{\acl{DATASET} Dataset}
\label{sec:dataset}
\begin{table}[t]
    \centering
    \caption{\acl{DATASET} is a large-scale sign language dataset containing extensive conversational data over multiple dynamic camera views and signers. The data covers a large number of environments, contexts, and topics, opening new research directions for the community. \textbf{Conv.}: dataset contains multi-party conversational signing. \textbf{Dynamic Camera}: footage involves moving or switching camera angles.}
    \label{tab:dataset_comparison}
    \begin{adjustbox}{max width=\textwidth}
    \begin{tabular}{lcccccccccc}
        \toprule
        \textbf{Dataset} & \textbf{Lang} & \textbf{Source} & \textbf{Env} & \textbf{\makecell{Signer\\Level}} & \textbf{Conv.} & \textbf{\makecell{Dynamic\\Camera}} & \textbf{\#Signers} & \textbf{\#Hours} & \textbf{\makecell{Text\\Vocab}} \\
        \midrule
        Phoenix-2014T~\cite{camgoz2018neural}  & DGS    & TV  & Studio & Interpreter        & \ding{55} & \ding{55} & 9    & 11    & 3K   \\
        How2Sign~\cite{duarte2021how2sign}       & ASL    & Lab & Studio & Interp./Native     & \ding{55} & \ding{55} & 11   & 79    & 16K  \\
        OpenASL~\cite{openasl}        & ASL    & Web & Varied & N/A                & \ding{55} & \ding{55} & 220  & 288   & 33K  \\
        YouTube-ASL~\cite{ytasl}    & ASL    & Web & Varied & N/A                & \ding{55} & \ding{55} & 2519 & 984   & 60K  \\
        YouTube-SL-25~\cite{tanzeryoutube}  & Varied & Web & Varied & N/A                & \ding{55} & \ding{55} & 3000 & 3207  & --   \\
        CSL-News~\cite{liuni}       & CSL    & TV  & Studio & Interpreter        & \ding{55} & \ding{55} & --   & 1985  & 5K   \\
        BOBSL~\cite{albanie2021bbc}          & BSL    & TV  & Studio & Interpreter        & \ding{55} & \ding{55} & 39   & 1467  & 77K  \\
        \midrule
        CSL-Daily~\cite{zhou2021improving}      & CSL    & Lab & Studio & Native             & \ding{55} & \ding{55} & 10   & 23    & 2K   \\
        BSL Corpus~\cite{bslcp}     & BSL    & Lab & Studio & Native             & \ding{51} & \ding{55} & 249  & 15$^{\dag}$ & -- \\
        MeineDGS~\cite{dgscorpus_3}       & DGS    & Lab & Studio & Native             & \ding{51} & \ding{55} & 330  & 50    & 18K  \\
        \midrule
        \textbf{\acl{DATASET}} & BSL & Web  & Varied & Native            & \ding{51} & \ding{51} & 2184 & 78.1  & 12K  \\
        \bottomrule
    \end{tabular}
    \end{adjustbox}
\end{table}

Progress in sign language modeling is fundamentally constrained by the quality of available data. Existing sign language datasets each leave distinct gaps for training generalisable models. Broadcast resources~\cite{camgoz2018neural, albanie2021bbc, liuni} offer substantial volume, but feature interpreter signing in static environments, carrying structural biases and artifacts from the source spoken language. Web-scraped collections~\cite{tanzeryoutube, ytasl} provide scale and visual variety, but lack verified signer profiles, linguistic quality control, and conversational interaction. Lab-recorded datasets~\cite{Schembri2013BuildingTB, dgscorpus_3, duarte2021how2sign, zhou2021improving} include rich linguistic content but are confined to fixed setups. Although some feature conversational interaction~\cite{Schembri2013BuildingTB, dgscorpus_3}, they remain visually sterile, failing to prepare models for real-world complexity.

To address these limitations, we introduce \acl{DATASET}, the first large-scale dataset of content natively produced in sign language. Created by Deaf signers, the dataset captures sign language in its most authentic, natural form. \acl{DATASET} consists of 44,759 atomic signing segments, sourced from 317 longer videos. This amounts to 78.1 hours of high-resolution video featuring an estimated 2,184 unique signers. Table.~\ref{tab:dataset_comparison} compares \acl{DATASET} to other sign language datasets.
A defining feature of \acl{DATASET} is its significant volume of conversational scenes involving two or more participants (approximately 40\% of the dataset). This natural interactivity, lacking in previous datasets, unlocks critical research directions that are impossible to explore with single-signer corpora. They include the analysis of multi-signer dynamics, turn-taking, signer diarisation, and the use of non-manual features as backchanneling and discourse markers. By providing authentic multi-party conversations, \acl{DATASET} pushes the field past translating isolated sentences towards natural sign language translation and expressive, realistic production.
To accelerate this research, we provide a comprehensive set of multi-modal annotations. For each video, we provide meta data including time-aligned English transcripts, extracted 2D human body keypoints, per-signer activity scores, depth maps, SMPL \cite{loper2015smpl} body, MANO \cite{romero2022embodied} hand, and FLAME \cite{FLAME:SiggraphAsia2017} face parameters. We also provide three data partitions that navigate different trade-offs between signer overlap and content diversity to allow for nuanced model evaluation. A more detailed description of the dataset construction is provided in the supplementary material.
\section{Methodology}
\label{sec:method}
\begin{figure}[t]
    \centering
    \includegraphics[width=\linewidth]{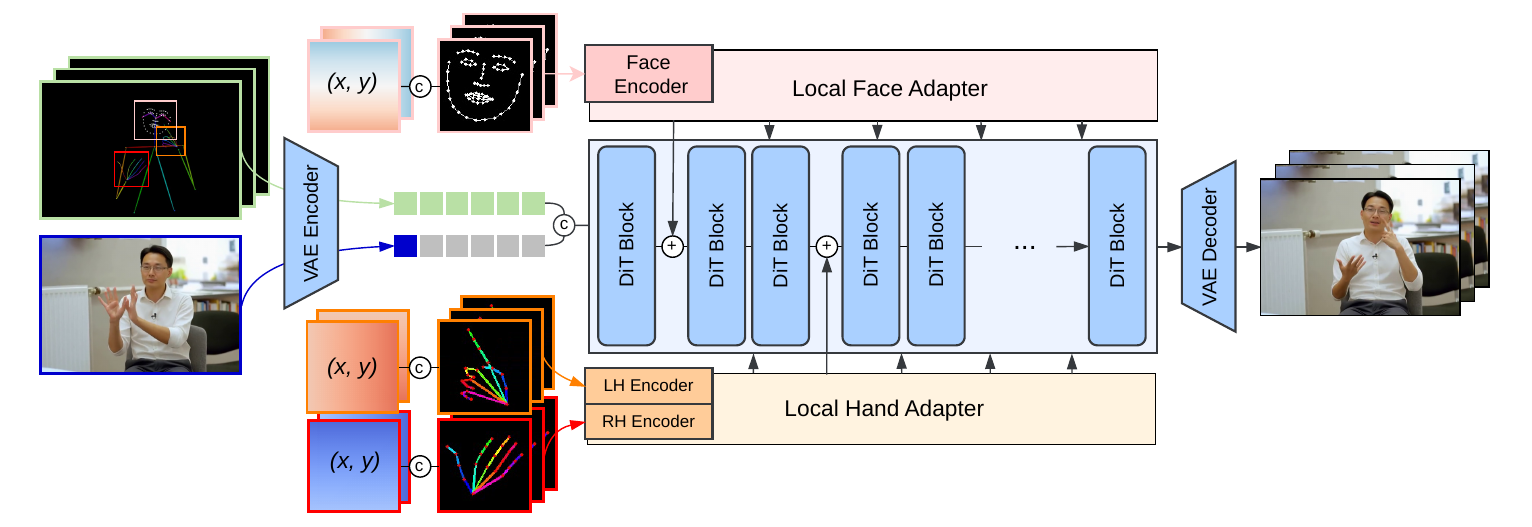}
    \caption{Architecture overview of SignRefine. A reference image and a full-body skeleton sequence guide the DiT backbone for video generation. Regional conditions for the face, left hand, and right hand are processed by separate encoders and injected into the DiT backbone through trainable local adapters, providing targeted refinement of the hand and face regions.}
    \label{fig:model_arch}
\end{figure}

Given a reference image defining signer appearance and a keypoint sequence specifying sign language motion, our goal is to generate a video depicting the signer faithfully reproducing the motion with precise hand articulations and facial expressions. Our system is built on Wan-Fun-Control-1.3B~\cite{wang2025wan}, a pretrained video DiT conditioned on skeleton sequences, selected for its strong generation quality and accessible size. While the full-body skeleton defines global body movements effectively, the model lacks precision for hands and face (regions that are spatially small yet linguistically critical). We introduce trainable local adapters that extract these regional conditions at higher resolution, encode them into compact motion features, and inject them into selected DiT blocks via masked cross-attention. The backbone remains frozen, preserving its generalised video priors while the adapters learn to steer the generation toward faithful and precise articulation. The overall architecture is illustrated in Fig.~\ref{fig:model_arch}.

\subsection{Regional Conditions with Spatial Grounding}

Signed communication happens through two primary articulatory channels: manual (hands) and non-manual (face, head and upper body). While these channels work cohesively to convey linguistic meaning, they present vastly different morphological features. The face requires modeling subtle, continuous deformations such as lip shapes and eyebrow movements, whereas the hands involve highly articulated, high-frequency structural changes. To account for this, we process the face, left hand, and right hand through three independent condition encoders, allowing each pathway to specialize its feature representation. Each encoder processes a regional keypoint sequence rendered as a $256 \times 256$ video into a sequence of motion tokens. A ResNet-style backbone first downsamples the spatial dimensions to a $16 \times 16$ feature grid. To capture motion dynamics, a pair of causal 1D convolutions then compresses the temporal dimension from T frames to T/4. This specific temporal compression aligns the regional condition with the temporal resolution of the backbone's latent space, producing motion vectors $\mathbf{M} \in \mathbb{R}^{T/4 \times S \times d}$ per region, where $S$ is the number of spatial latent patches and $d$ the hidden dimension size.

A fundamental challenge in this regional approach is spatial ambiguity: once extracted, a local crop loses the spatial context of its original location within the video. Because the adapter must project these features back into the correct location of the full frame, the encoder needs global positional awareness. We resolve this by injecting positional information into the input channels of the crop, inspired by the CoordConv approach \cite{liu2018intriguing}. For every pixel $(i, j)$ in the crop, we compute its absolute normalized coordinate relative to the original full frame:
\begin{equation}
    x_{i,j} = \frac{2(x_1 + (x_2 - x_1) \cdot j / W)}{W_\text{orig}} - 1, \quad y_{i,j} = \frac{2(y_1 + (y_2 - y_1) \cdot i / H)}{H_\text{orig}} - 1,
\end{equation}
where $(x_1, y_1, x_2, y_2)$ are the bounding box corners in pixel coordinates and $W_\text{orig}, H_\text{orig}$ are the original frame dimensions. This maps each crop pixel to its position in the full frame, normalized to $[-1, 1]$. By providing these dynamic coordinate maps alongside the RGB channels, we directly embed the crop-to-frame geometric mapping into the regional features.

\subsection{Localized Feature Injection}

The extracted motion vectors must be integrated into the DiT backbone to provide precise and localized control over the articulators without corrupting the background. We achieve this through cross-attention blocks injected at regular intervals across the DiT. The backbone consists of 30 transformer blocks, the face adapter injects cross-attention at blocks $\{0, 6, 12, 18, 24\}$ and the hand adapter at blocks $\{2, 8, 14, 20, 26\}$. This offset between face and hand injection prevents any single DiT block from receiving multiple adapter residuals simultaneously, distributing the conditioning load across the network. Furthermore, the left and right hand conditions are concatenated before injection to better resolve inter-hand dependencies and occlusions.

Because the DiT operates in a compressed latent space, each token in the hidden state $\mathbf{x} \in \mathbb{R}^{S \times d}$ maintains a predictable spatial correspondence to a specific area in the original pixel space. This allows us to map the regional bounding box coordinates onto the coarse latent grid, producing a binary spatial mask $\mathbf{m} \in \{0, 1\}^S$ that separates the target articulator (foreground) from the rest of the frame (background). To reinforce this spatial isolation during feature injection, we append a trainable sink token $\mathbf{s} \in \mathbb{R}^{1 \times d}$ to the sequence of motion vectors $\mathbf{M}$. We compute queries from the DiT hidden states and keys/values from the extended motion sequence:
\begin{equation}
    \mathbf{Q} = W_Q \cdot \text{LN}(\mathbf{x}), \quad \mathbf{K}, \mathbf{V} = W_{KV} \cdot \text{LN}([\mathbf{M}; \mathbf{s}])
\end{equation}
$W_Q$ and $W_{KV}$ are the learned linear projection weight matrices for the queries, keys and values, respectively. $\text{LN}(\cdot)$ denotes the Layer Normalization operation. Attention is then routed using the spatial mask:
\begin{equation}
    A_{ij} = \begin{cases}
    \text{softmax}(\mathbf{q}_i \cdot \mathbf{k}_j / \sqrt{d}) & \text{if } m_i = 1 \text{ and } j \leq N \text{ (foreground} \to \text{content)} \\
    \text{softmax}(\mathbf{q}_i \cdot \mathbf{k}_\text{sink} / \sqrt{d}) & \text{if } m_i = 0 \text{ (background} \to \text{sink)}
    \end{cases},
\end{equation}
where $A_{ij}$ is the computed attention weight mapping the $i$-th query patch $\mathbf{q}_i$ to the $j$-th key token $\mathbf{k}_j$; $N = |\mathbf{M}|$ is the length of the motion-vector sequence, so the keys at indices $j \leq N$ are the projected motion tokens, and the key at index $N{+}1$, denoted $\mathbf{k}_\text{sink}$, is the projection of the appended sink token $\mathbf{s}$; $m_i \in \{0, 1\}$ represents the binary spatial mask value for the $i$-th latent patch: $m_i = 1$ indicates the patch falls inside the articulator's bounding box (foreground), and $m_i = 0$ indicates its belonging to the background; $\sqrt{d}$ is the scaling factor to stabilize gradients.

This mechanism ensures that the foreground latent patches attend exclusively to the high-resolution regional features, while the background patches are routed to the sink token. The sink token acts as a null target, safely absorbing background queries and preventing the adapter from bleeding hand or face features into unrelated areas.

\subsection{Adapter Training}

The output projection layers of the regional adapters are initialized with near-zero values \cite{zhang2023adding} to ensure training stability on top of the frozen backbone. The training objective is the mean squared error (MSE) on the predicted noise with spatially non-uniform weighting that emphasizes the regions the adapter is designed to improve. Using the same bounding box masks available from the conditioning pipeline, we partition each frame's latents into three regions, hands, face, and background, and combine their mean errors into
a single objective:
\begin{equation}
    \mathcal{L} = \frac{w_h N_h \mathcal{L}_h + w_f N_f \mathcal{L}_f + w_b N_b \mathcal{L}_b}{w_h N_h + w_f N_f + w_b N_b},
\end{equation}
where $\mathcal{L}_h, \mathcal{L}_f, \mathcal{L}_b$ are the MSE within the hand, face, and background regions, $N_h, N_f, N_b$ are the numbers of latent elements they contain, and $w_h, w_f, w_b$ are their weights. This is equivalent to assigning every latent element the weight of its region and taking a single per-element weighted average, so each region's influence on the gradient is proportional to both its weight and its area. The background region is defined as the complement of the hand and face masks, so the body (torso and arms outside the articulator boxes) is absorbed into the background term rather than receiving a dedicated weight. In overlapping regions, hands take priority over face, reflecting their common relative position in sign language.

We set $w_h = w_f = 10, w_b = 1$ in our experiments, motivated by the area imbalance. Although the hands and face regions are critical for sign comprehension, at the latent resolution they occupy only $\sim$15\% of each frame on average, so under uniform weighting they would receive a correspondingly small share of the training gradient. The chosen weights raise the articulators' share of the gradient, prioritizing them, while retaining roughly a third of the gradient on the background to preserve global coherence such as body pose and identity.
\section{Experiments}
\label{sec:experiments}
\subsection{Implementation Details}

We use Wan2.1-1.3B-Fun-Control \cite{wang2025wan} DiT as the frozen backbone, which includes a T5 text encoder, a CLIP image encoder for reference image conditioning, and a 3D VAE for encoding/decoding video latents. As varying the text prompt has a negligible effect on generation quality, we cache the latents for a generic text prompt and load them once to improve training efficiency. 

\noindent\textbf{Adapter architecture} ~ Each regional condition encoder uses a ResNet-style convolutional backbone with GroupNorm \cite{wu2018group} and SiLU \cite{elfwing2018sigmoid} activations, downsampling the $256 \times 256$ input video into a $16 \times 16$ spatial grid. Temporal downsampling is performed by a stack of causal 1D convolutions, compressing $T$ frames to $T/4$. The final linear layer projects the features from 512 to 1536 dimensions, which matches the hidden size of the Wan backbone. There are two separate adapters, one for face and one for hand regions, each including 6 cross-attention blocks. Each block uses 12 heads with head dimension 128 and RMSNorm \cite{zhang2019root} on queries and keys.

\noindent\textbf{Training} ~ We train the model on $\sim40,000$ video clips from the \acl{DATASET} train partition for 30,000 iterations. We use Adam optimizer \cite{diederik2014adam} with learning rate $1e-4$, gradient accumulation over 8 iterations, the batch size of 1 per-GPU. The model is trained on 4 NVIDIA GH200 GPUs in mixed precision mode.

\begin{figure*}[t]
    \centering
    \includegraphics[width=\textwidth]{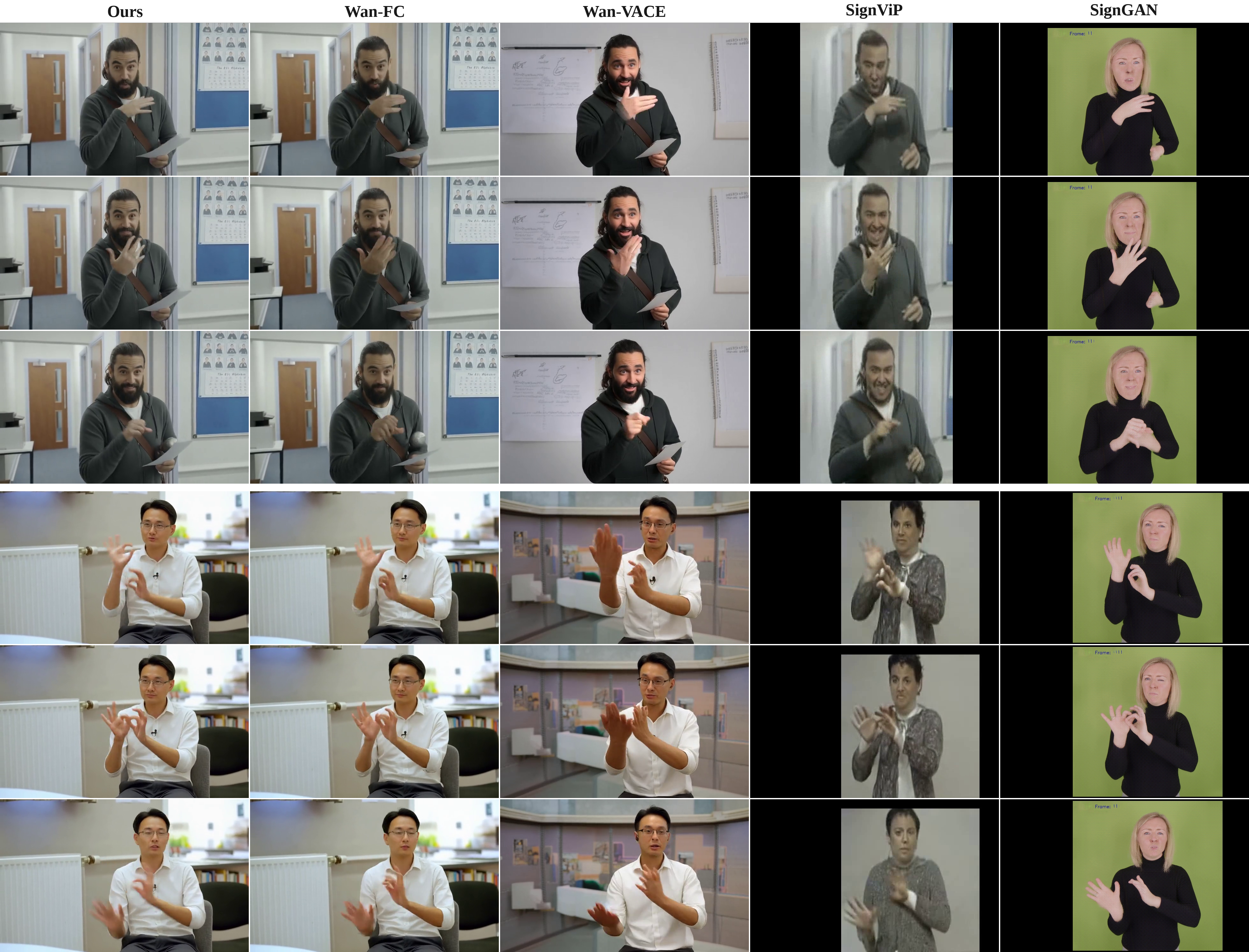}
    \caption{Qualitative comparison of results from different models. From left to right: Ours, Wan-FC, Wan-VACE, SignViP, SignGAN.}
    \label{fig:qualitative_examples}
\end{figure*}

\subsection{Quantitative Evaluation}

The evaluation is performed on the \acl{DATASET} appearance-independent test set, \acl{PHOENIX} \cite{camgoz2018neural} and \acl{BSLCP} \cite{bslcp} to assess the generalization of the models. We randomly choose 100 videos from each dataset to keep the inference time manageable. As the main goal of this work is to improve the quality of the hand and face regions, we concentrate our evaluation on regional metrics. We use LPIPS \cite{zhang2018unreasonable} with the VGG \cite{Simonyan15} backbone and SSIM \cite{wang2004image} to assess perceptual quality. MPJPE \cite{ionescu2013human3} is chosen as a measure of keypoint accuracy and structural precision. For its evaluation, the 133 full-body keypoints from both ground-truth (GT) and generated videos are extracted with RTMPose-X \cite{jiang2023rtmpose}. All regional crops are extracted using the bounding boxes derived from the GT keypoints.

We compare our approach with several state-of-the-art models in video generation. Wan2.1-1.3B-Fun-Control (Wan-FC) and Wan2.1-1.3B-VACE (Wan-VACE) \cite{jiang2025vace} are general DiT video generation models that are driven with a full-body skeleton condition. SignViP \cite{wang2025advanced} is a multi-step video model, specifically trained on sign language data. It encodes the full body skeleton and 3D hand mesh renders as latent tokens, which are then used to condition the video generator. For a fair evaluation, we construct the tokens from the conditions extracted from the GT videos. We use the model checkpoint trained on \acl{PHOENIX} as it is the only checkpoint made available by the authors. As there is no open-source implementation of SignGAN \cite{saunders2020everybody}, we use the implementation and the checkpoint obtained directly from the authors. The obtained model was trained on one fixed appearance and does not accept visual conditioning; therefore, the results will not follow the GT visually. However, as we perform the evaluation on regional crops, the influence of the overall appearance is reduced, and the comparison to other models is fair.

The results for all datasets are presented in Table~\ref{tab:main_results} and Fig.~\ref{fig:qualitative_examples} shows qualitative examples generated by different models. Our method consistently achieves the strongest performance across all three anatomical regions and all three datasets, with particularly pronounced gains on hand regions. The improvement over the strongest baseline, Wan-FC, is most evident in hand fidelity on \acl{DATASET}, where our model shows roughly a 30\% reduction in MPJPE. Face precision also improves consistently across datasets, by up to 20\% on \acl{PHOENIX} and 15\% on \acl{BSLCP}. The Wan-VACE model, as well as sign-specific SignViP and SignGAN, consistently underperform in all testing scenarios.

A notable pattern emerges when examining the generalization behavior of the prior methods. SignViP, trained exclusively on \acl{PHOENIX}, suffers a severe performance collapse on \acl{DATASET} and \acl{BSLCP}, with hand MPJPE roughly doubling compared to its in-domain results. This indicates that its learned representations are tightly coupled to the visual statistics of its training corpus. Similarly, SignGAN, trained on controlled studio recordings, exhibits the weakest overall performance, particularly on the \acl{DATASET} data, underscoring the well-known domain gap between laboratory and in-the-wild signing conditions. In contrast, our method maintains stable performance across all datasets without any dataset-specific adaptation, demonstrating that the proposed architecture generalizes robustly across diverse signing environments and visual domains.

\begin{table}[t]
    \scriptsize
    \centering
    \caption{Quantitative comparison on \acl{DATASET} (top), \acl{PHOENIX} (middle) and \acl{BSLCP} (bottom) datasets. Best results are \textbf{bolded}, and second-best are \underline{underlined}.}
    \label{tab:main_results}
    \setlength{\tabcolsep}{4pt}
\begin{adjustbox}{width=\linewidth}
\begin{tabular}{@{}l ccc ccc ccc@{}}
    \toprule
    \multirow{2}{*}{\textbf{Method}} & \multicolumn{3}{c}{\textbf{Face}} & \multicolumn{3}{c}{\textbf{Right Hand (RH)}} & \multicolumn{3}{c}{\textbf{Left Hand (LH)}} \\
    \cmidrule(lr){2-4} \cmidrule(lr){5-7} \cmidrule(l){8-10}
    & MPJPE$\downarrow$ & LPIPS$\downarrow$ & SSIM$\uparrow$ & MPJPE$\downarrow$ & LPIPS$\downarrow$ & SSIM$\uparrow$ & MPJPE$\downarrow$ & LPIPS$\downarrow$ & SSIM$\uparrow$ \\
    \midrule
    \rowcolor{gray!10}\multicolumn{10}{c}{\textit{\acl{DATASET} Dataset}} \\
    SignGAN \cite{saunders2020everybody}     & 10.424 & 0.564 & 0.302 & 24.774 & 0.627 & 0.388 & 21.852 & 0.620 & 0.384 \\
    SignViP \cite{wang2025advanced}    & 10.576 & 0.534 & 0.356 & 23.957 & 0.581 & 0.428 & 24.383 & 0.588 & 0.424 \\
    Wan-VACE  \cite{jiang2025vace} & 4.067  & 0.422 & 0.367 & 15.904 & 0.525 & 0.380 & 15.431 & 0.503 & 0.407 \\
    Wan-FC \cite{wang2025wan}  & \underline{1.835} & \underline{0.197} & \underline{0.713} & \underline{9.565} & \underline{0.364} & \underline{0.583} & \underline{9.341} & \underline{0.349} & \underline{0.597} \\
    \textbf{Ours} & \textbf{1.770} & \textbf{0.194} & \textbf{0.714} & \textbf{6.602} & \textbf{0.330} & \textbf{0.629} & \textbf{6.638} & \textbf{0.321} & \textbf{0.640} \\
    \midrule
    \rowcolor{gray!10}\multicolumn{10}{c}{\textit{\acl{PHOENIX} Dataset}} \\
    SignGAN \cite{saunders2020everybody}      & 5.457 & 0.585 & 0.343 & 28.507 & 0.707 & 0.469 & 25.373 & 0.657 & 0.470 \\
    SignViP \cite{wang2025advanced}     & 7.194 & 0.456 & 0.276 & 10.345 & 0.534 & 0.425 & 14.405 & 0.566 & 0.376 \\
    Wan-VACE \cite{jiang2025vace} & 2.162 & 0.383 & 0.413 & 11.717 & 0.551 & 0.367 & 13.214 & 0.546 & 0.299 \\
    Wan-FC \cite{wang2025wan}  & \underline{1.151} & \underline{0.223} & \underline{0.688} & \underline{5.754} & \underline{0.402} & \underline{0.629} & \underline{5.719} & \underline{0.374} & \underline{0.642} \\
    \textbf{Ours} & \textbf{0.916} & \textbf{0.214} & \textbf{0.701} & \textbf{4.582} & \textbf{0.386} & \textbf{0.660} & \textbf{5.101} & \textbf{0.366} & \textbf{0.668} \\
    \midrule
    \rowcolor{gray!10}\multicolumn{10}{c}{\textit{\acl{BSLCP} Dataset}} \\
    SignGAN \cite{saunders2020everybody}     & 9.415 & 0.634 & 0.330 & 22.726 & 0.720 & 0.406 & 16.858 & 0.689 & 0.390 \\
    SignViP \cite{wang2025advanced}    & 28.222 & 0.626 & 0.289 & 23.085 & 0.611 & 0.381 & 22.111 & 0.593 & 0.372 \\
    Wan-VACE  \cite{jiang2025vace} & 3.172  & 0.436 & 0.380 & 17.524 & 0.567 & 0.351 & 15.385 & 0.557 & 0.338 \\
    Wan-FC \cite{wang2025wan}  & \underline{1.790} & \textbf{0.221} & \underline{0.685} & \underline{8.232} & \underline{0.389} & \underline{0.559} & \underline{7.002} & \underline{0.373} & \underline{0.563} \\
    \textbf{Ours} & \textbf{1.524} & \underline{0.222} & \textbf{0.690} & \textbf{6.314} & \textbf{0.365} & \textbf{0.608} & \textbf{5.406} & \textbf{0.353} & \textbf{0.607} \\
    \bottomrule
\end{tabular}
\end{adjustbox}
\end{table}

\subsection{User Study}

\noindent\textbf{Protocol} ~ To strengthen the quality and understandability evaluation of the models, we conduct a perceptual study with 15 participants of varying British Sign Language (BSL) proficiency, from basic to advanced. Each participant completed 17 questions across two question types. In the first, \emph{Visual Ranking}, participants viewed four videos generated by different models from the same conditioning input and ranked them by naturalness of signing and quality of hand and facial features. These questions include a combination of samples from \acl{DATASET} and \acl{PHOENIX}. Although BSL signers can not directly evaluate the linguistic accuracy of DGS (German Sign Language) in \acl{PHOENIX} videos, they can still assess and compare their visual fidelity. The second type, \emph{Text Reference Ranking}, is only shown to participants with intermediate or higher BSL proficiency. Given a reference English text sentence, participants are asked to rank four BSL video renditions by semantic fidelity. To ensure a fair comparison among five models using a four-slot display, each question excludes one model according to a balanced rotation schedule that guarantees approximately equal representation. Model-to-label assignments per question and the question order are randomised to prevent positional bias. We analyse the collected preferences through three complementary metrics: (1) \emph{Mean Rank} aggregates each model's average position across all participant-question instances, with pairwise significance against our method assessed via Wilcoxon signed-rank tests on co-occurring model pairs, corrected for multiple comparisons using the Holm-Bonferroni procedure; (2) \emph{Rank-1 Win Rate} measures how often each model was placed first, reported with the Wilson score confidence intervals against a 25\% chance baseline; (3) \emph{Head-to-Head Preference} matrix captures the proportion of co-occurrences in which each model was preferred over every other, indicating detailed dominance relationships beyond aggregate rank.

\begin{table*}[b]
    \centering
    \scriptsize
    \setlength{\tabcolsep}{2pt} 
    
    \begin{minipage}{0.52\textwidth}
        \centering
        \caption{\textbf{Perceptual Quality study results.} Mean rank assigned by participants to each model (1 = best, 5 = worst), reported overall and per evaluation category.}
        \label{tab:ranking_results}
        \begin{tabular}{l ccccc} 
        \toprule
        \textbf{Category} & \textbf{Ours} & \begin{tabular}{@{}c@{}}\textbf{Wan}\\\textbf{-FC}\end{tabular} & \begin{tabular}{@{}c@{}}\textbf{Wan}\\\textbf{-VACE}\end{tabular} & \begin{tabular}{@{}c@{}}\textbf{Sign}\\\textbf{ViP}\end{tabular} & \begin{tabular}{@{}c@{}}\textbf{Sign}\\\textbf{GAN}\end{tabular} \\
        \midrule
        Overall $\downarrow$        & \textbf{1.22} & \underline{1.89} & 2.46 & 3.56 & 3.41 \\
        \acl{DATASET} $\downarrow$        & \textbf{1.16} & \underline{1.95} & 2.41 & 3.81 & 3.23 \\
        Phoenix14T $\downarrow$     & \textbf{1.41} & \underline{1.89} & 2.60 & 2.65 & 3.95 \\
        Text Ref. $\downarrow$      & \textbf{1.16} & \underline{1.63} & 2.44 & 3.76 & 3.22 \\
        \bottomrule
        \end{tabular}
    \end{minipage}%
    \hfill 
    \begin{minipage}{0.43\textwidth}
        \centering
        \caption{\textbf{Comprehension study results.} Free-text translations from fluent signers, scored against ground-truth references with four complementary metrics.}
        \label{tab:comprehension}
        \begin{tabular}{l cc}
        \toprule
        \textbf{Metric} & \textbf{Ours} & \textbf{Wan-FC} \\
        \midrule
        Keyword Recall $\uparrow$       & \textbf{0.54} & 0.30 \\
        BLEURT $\uparrow$               & \textbf{0.31} & 0.19 \\
        LLM adequacy $\uparrow$         & \textbf{2.07} & 1.50 \\
        Failure Rate $\downarrow$       & \textbf{0.17} & 0.27 \\
        \bottomrule
        \end{tabular}
    \end{minipage}
\end{table*}

\begin{figure}[t]
    \centering
    \includegraphics[width=\linewidth]{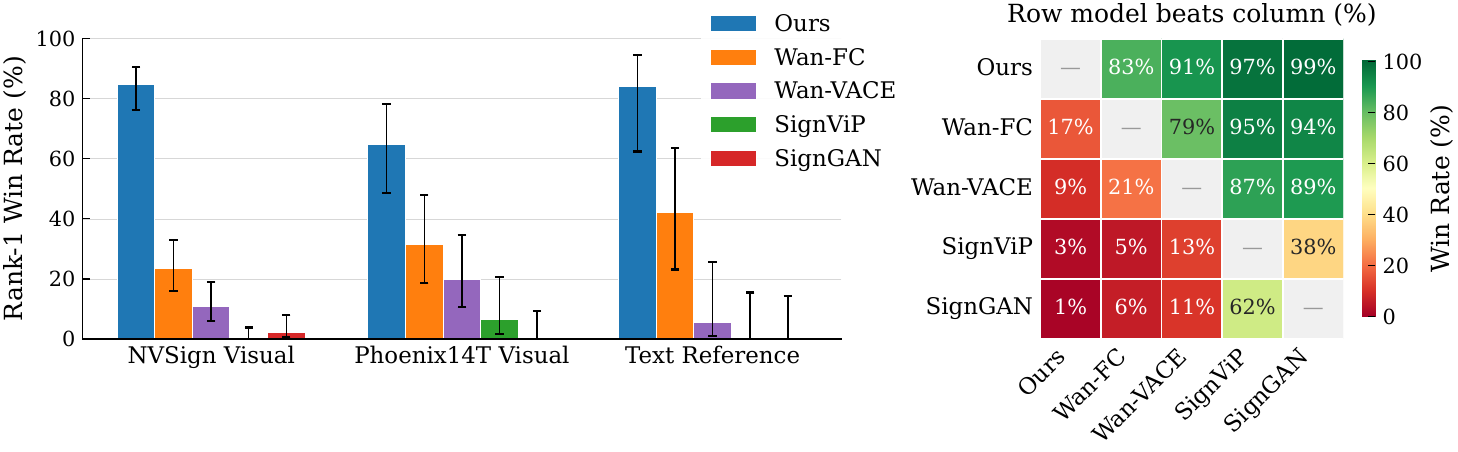}
    \caption{\textbf{Perceptual Quality study results.} Left: Rank-1 win rates with confidence intervals reported for each question category. Right: Head-to-head preference matrix, aggregated across all three categories, showing the percentage of co-occurrences in which the row model is preferred over the column model.}
    \label{fig:user_study_results}
\end{figure}

\noindent\textbf{Results} ~ We collect approximately 200 observations per model under the balanced exclusion schedule. The results are presented in Table~\ref{tab:ranking_results} and Fig.~\ref{fig:user_study_results}. Our method achieves the lowest mean rank overall and across both question types, with every baseline performing worse under Holm-corrected Wilcoxon tests ($p<0.001$). Wan-FC emerges as the second-strongest method, followed by Wan-VACE. SignGAN and SignViP occupy the bottom two positions with mean ranks above 3.4. The gap between our method and Wan-FC is narrowest on \acl{PHOENIX}, where the low video resolution of $210 \times 260$ reduces the quality ceiling for highly capable Wan models, yet the difference remains significant. Rank-1 win rates reinforce this ordering: our method is ranked first in the vast majority of \acl{DATASET} Visual and Text Reference trials, and in roughly two-thirds of \acl{PHOENIX} Visual trials. Wan-FC is the only baseline that consistently exceeds chance, while SignViP and SignGAN are almost never ranked first. The head-to-head preference matrix from Fig.~\ref{fig:user_study_results} reveals that our method is preferred over every individual baseline in more than 83\% of direct comparisons. A clear preference hierarchy emerges from the user study responses: Ours $>$ Wan-FC $>$ Wan-VACE $>$ SignGAN $\approx$ SignViP, which corresponds to the ranking obtained during the quantitative evaluation.

\noindent\textbf{Comprehension} ~ To determine whether the observed improvements in visual quality of our model's outputs translated into enhanced linguistic intelligibility, we conducted an independent comprehension study with six fluent BSL signers. We selected ten short BSL clips from the YouTube-SL-25 dataset \cite{tanzeryoutube}, each having a ground-truth English reference translation, and rendered them with our model and the best-performing baseline: Wan-FC. Participants viewed a randomly chosen rendition of every clip, blind to its source model, and typed an English translation of the perceived content or flagged the clip as incomprehensible, producing 60 responses, 30 per model. The clip order was randomized to suppress narrative priming. Because free-text responses are inherently noisy, often containing paraphrases, telegraphic phrasing, and filler words, we evaluate the data using four complementary metrics to maximize robustness: (1) \emph{Failure Rate} -- fraction of incomprehensible videos; (2) \emph{Keyword Recall} -- fraction of each clip's salient facts such as names, figures, dates recovered; (3) \emph{BLEURT-20} \cite{sellam-etal-2020-bleurt}, a learned metric calibrated to mimic human judgments of translation quality; and (4) an \emph{LLM Adequacy Score} produced by a state-of-the-art LLM (Claude Opus 4.8), prompted to evaluate how much of each reference clip's meaning is conveyed by translations on a scale of 0 to 4. Incomprehensible responses score zero on the continuous metrics, and the significance of each metric is assessed with a clip-paired Wilcoxon signed-rank test across the ten clips, which controls for clip difficulty. As reported in Table~\ref{tab:comprehension}, our model is more comprehensible on every metric: viewers recover more of each video's key content, their translations are judged closer in meaning to the reference by both a learned metric and an LLM rater, and they abandon fewer clips as unintelligible. This advantage is statistically significant for Keyword Recall and BLEURT ($p{<}0.05$), while LLM adequacy and Failure Rate exhibit the same effect as a consistent trend. Although fluent signers are a scarce population and our panel is correspondingly small, the convergent evidence across complementary, methodologically distinct metrics provides confident support that our approach produces measurably more understandable sign language video.

\subsection{Ablation Study}

\begin{table}[b]
\scriptsize
    \centering
    \caption{Ablation study of different adapter configurations and components.}
    \label{tab:ablation_study}

    \begin{tabular}{@{}l cc cc cc@{}}
    \toprule
    \multirow{2}{*}{\textbf{Method}} & \multicolumn{2}{c}{\textbf{Face}} & \multicolumn{2}{c}{\textbf{LH (Left Hand)}} & \multicolumn{2}{c}{\textbf{RH (Right Hand)}} \\
    \cmidrule(lr){2-3} \cmidrule(lr){4-5} \cmidrule(l){6-7}
    & MPJPE$\downarrow$ & LPIPS$\downarrow$ & MPJPE$\downarrow$ & LPIPS$\downarrow$ & MPJPE$\downarrow$ & LPIPS$\downarrow$ \\
    \midrule
    Dense conditions               & \textbf{1.692} & \textbf{0.192} & 6.665          & \textbf{0.319} & 6.713          & \textbf{0.329} \\
    Sparse conditions        & 1.770          & 0.194          & \textbf{6.638} & 0.321          & \textbf{6.602} & 0.330          \\
    Sparse w/o attention masking  & 1.949          & 0.214          & 7.309          & 0.353          & 7.270          & 0.363          \\
    Sparse w/o regional coords    & 1.847          & 0.202          & 6.926          & 0.335          & 6.888          & 0.344          \\
    \bottomrule
    \end{tabular}
\end{table}

We conduct an ablation study on some of the key components of our proposed adapters, which include regional attention masking and CoordConv coordinates. We also experiment with providing dense conditions to the adapters, rendered face normals and 3D MANO meshes, instead of regional keypoint renders. The results are provided in Table~\ref{tab:ablation_study}. Removing masked attention degrades all metrics, with face MPJPE increasing from 1.770 to 1.949 (+10.1\%) and hand MPJPE rising by a similar margin (e.g. LH from 6.638 to 7.309). Masked attention focuses the cross-attention on relevant spatial regions, preventing the adapter from concentrating on uninformative background tokens. Removing the CoordConv input encoding yields a smaller but consistent degradation across all metrics (+4.3\%), confirming that the absolute spatial position helps the model disambiguate between regions.

When comparing sparse and dense conditions, the dense adapter achieves marginally better face metrics, possibly owing to the richer geometric detail in rendered face normals, but yields no improvement for hands. We attribute this to two factors. Firstly, our region adapters get conditions in $256\times256$ resolution, which significantly exceeds the area the hands typically occupy in original $832\times480$ frames. We find that this magnification, rather than condition density, is the primary driver of hand quality: at this resolution, sparse keypoints already provide sufficient structural guidance. Secondly, the pretrained 1.3B-parameter DiT backbone encodes a strong prior over hand appearance and articulation, allowing it to synthesize plausible details from coarse positional cues alone. The additional surface geometry in MANO renders seems largely redundant in this case. Furthermore, sparse keypoints more closely resemble the skeleton-based control signals the backbone was designed to process and are less susceptible to the structured noise present in predicted MANO meshes. Given comparable hand performance, we adopt the sparse adapter as our default to eliminate the inference-time dependency on MANO and face normal prediction and rendering.
\section{Limitations}
\label{sec:limitations}
Although our proposed approach produces impressive visual results, it inherits several limitations. Firstly, video diffusion models remain computationally expensive: generating an 81-frame clip takes minutes even on high-end GPUs, precluding real-time applications. 
Secondly, our model reproduces the motion blur present in the training data, which is typically filmed at conventional frame rates (25 fps). Fast movements are prevalent in sign languages, that results in the model occasionally smearing fine hand details during rapid transitions.
Finally, both our conditioning pipeline and keypoint-based evaluation rely on off-the-shelf pose estimators such as RTMPose, WiLoR \cite{potamias2025wilor} \etc, whose errors propagate into the model. Noisy keypoints in the input condition can mislead generation and reduce the quality. Additionally, inaccuracies in the predicted keypoints frames add noise to the reference-based evaluation metrics, potentially obscuring the true performance differences between methods.
\section{Conclusion}
\label{sec:conclusion}
In this work, we presented SignRefine, a video generation model that adapts a pretrained Diffusion Transformer for sign language via local adapters with spatial grounding. By injecting the high-resolution regional conditions of the hands and face into the frozen backbone through masked cross-attention, our approach corrects the articulation failures of general-purpose video models without sacrificing their quality and generalisability. Furthermore, we introduced \acl{DATASET}, a large-scale natively-signed high resolution dataset, providing diverse signer appearances, dynamic visual conditions, and multi-party conversational scenes absent from all prior corpora. SignRefine achieves substantial improvement in hand pose precision over the strongest baseline and is preferred by sign language users in over 80\% of pairwise comparisons. This work demonstrates that foundational video models can be effectively used for sign language, opening new possibilities for high-fidelity, generalizable sign language content creation.

\section*{Acknowledgements}
{\sloppy
This work was supported by EPSRC grant APP24554 (SignGPT-EP/Z535370/1), EPSRC grant APP78083 (UMCS UKRI3927) and through funding from Google.org via the AI for Global Goals scheme.
The authors acknowledge the use of Isambard-AI National AI Research Resource (AIRR) funded by UK DSIT via UKRI and STFC [ST/AIRR/I-A-I/1023].
This work reflects only the authors' views and the funders are not responsible for any use that may be made of the information it contains.\par}

%
%
\bibliographystyle{splncs04}
\bibliography{main}
\end{document}